\documentclass[conference]{IEEEtran}
\IEEEoverridecommandlockouts

\usepackage[table]{xcolor} 
\usepackage{pgfplots}
\usepackage{graphicx, float}
\usepackage{booktabs,colortbl}
\usepackage{pdflscape}
\usepackage{amsfonts}
\usepackage{placeins}
\usepackage{subcaption}
\usepackage{mathtools,textcomp}
\usepackage{pgfplotstable}
\usepackage{nccmath}
\usepackage{amsmath,amssymb}
\usepackage{longtable,multirow}
\usepackage{hyperref}
\usepackage{xurl}
\usepackage[capitalize]{cleveref}
\def\BibTeX{{\rm B\kern-.05em{\sc i\kern-.025em b}\kern-.08em
    T\kern-.1667em\lower.7ex\hbox{E}\kern-.125emX}}

\newcommand{\degC}{{$^\circ$C }} 

\usepgfplotslibrary{fillbetween,statistics}
\usetikzlibrary{patterns,pgfplots.groupplots}
\pgfplotsset{compat=1.5}

\usetikzlibrary{external}
\definecolor{Blue}{RGB}{13,133,181}
\definecolor{Green}{RGB}{130,208,101}
\definecolor{Orange}{RGB}{209,106,19}

\makeatletter

\newcommand*{\citeSup}[2][]{%
  \begingroup
  \let\NAT@mbox=\mbox
  \let\@cite\NAT@citesuper
  \let\NAT@space\relax
  \renewcommand\NAT@open{[}%
  \renewcommand\NAT@close{]}%
  \cite[#1]{#2}%
  \endgroup
}
\makeatother

\def\BibTeX{{\rm B\kern-.05em{\sc i\kern-.025em b}\kern-.08em
    T\kern-.1667em\lower.7ex\hbox{E}\kern-.125emX}}

\begin{document}

\title{Adaptive-Sensorless Monitoring of Shipping Containers}

\author{
    \IEEEauthorblockN{
        Lingqing Shen\IEEEauthorrefmark{1}\IEEEauthorrefmark{2}\hspace{2em}
        Chi Heem Wong\IEEEauthorrefmark{1}\IEEEauthorrefmark{3}\hspace{2em}
        Misaki Mito\IEEEauthorrefmark{3}\hspace{2em}
        Arnab Chakrabarti\IEEEauthorrefmark{3}
    }
    \IEEEauthorblockA{
        \IEEEauthorrefmark{1} Equal contribution. Work done during Lingqing’s internship at Hitachi America \\
        \IEEEauthorrefmark{2}\textit{Tepper School of Business, Carnegie Mellon University, Pittsburgh, United States}\\
        \IEEEauthorrefmark{3}\textit{R\&D Division, Hitachi America, Ltd., Santa Clara, United States}\\
    }
}

\maketitle

\begin{abstract}
\noindent%
Monitoring the internal temperature and humidity of shipping containers is essential to preventing quality degradation during cargo transportation.
Sensorless monitoring---machine learning models that predict the internal conditions of the containers using exogenous factors---shows promise as an alternative to monitoring using sensors.
However, it does not incorporate telemetry information and correct for systematic errors, causing the predictions to differ significantly from the live data and confusing the users.
In this paper, we introduce the residual correction method, a general framework for correcting for systematic biases in sensorless models after observing live telemetry data.
We call this class of models ``adaptive-sensorless'' monitoring.
We train and evaluate adaptive-sensorless models on the 3.48 million data points---the largest dataset of container sensor readings ever used in academic research---and show that they produce consistent improvements over the baseline sensorless models.
When evaluated on the holdout set of the simulated data, they achieve average mean absolute errors (MAEs) of 2.24 $\sim$ 2.31\degC (vs 2.43\degC by sensorless) for temperature and 5.72 $\sim$ 7.09\% for relative humidity (vs 7.99\% by sensorless) and average root mean-squared errors (RMSEs) of 3.19 $\sim$ 3.26\degC for temperature (vs 3.38\degC by sensorless) and 7.70 $\sim$ 9.12\% for relative humidity (vs 10.0\% by sensorless).
Adaptive-sensorless models enable more accurate cargo monitoring, early risk detection, and less dependence on full connectivity in global shipping. 
\end{abstract}

\begin{IEEEkeywords}
Adaptive monitoring, Supply chain, Artificial Intelligence, Big Data, Analytics
\end{IEEEkeywords}

\section{Introduction}
\label{sec:intro}

Shipping containers, also known as intermodal containers or freight containers, transport a large fraction of global trade~\cite{unctd2022review}.
A significant portion of this cargo suffers quality degradation during transport, often due to mold in perishables (e.g., grain or nuts) and packaged products (e.g., bagged milk powder) or corrosion in metals and metallic packaging (e.g., canned goods). 
Given the large volume of global trade, losses reach billions of dollars annually.

The rates of quality degradation are functions of temperature and humidity (\cite{viitanen2007improved,viitanen2015mold}) and monitoring these variables is instrumental in predicting and ultimately preventing damage.
Current risk mitigation strategies use sensors placed inside shipping containers to record and transmit the data for analysis.
However, such sensor-based solutions are expensive owing to the cost of purchasing sensing electronics, the operational cost of sustaining multimodal telecommunication as the containers travel around the world, and the labor cost of maintenance, repair, and battery replacement.

To address these limitations, Mito et al.~\cite{mito2023sensorless} developed \emph{sensorless} monitoring, a machine learning method using weather data to estimate the internal temperature and relative humidity of the containers.
Although resourceful, it suffers from a couple of deficiencies.
First, it trains on only 85 routes across 7 countries, which raises questions about the generalizability of its model to all locations and climates.
Second and more importantly, sensorless monitoring uses only exogenous data and does not incorporate live sensor data.
This causes sensorless monitoring to deviate from the live sensor readings---which are often available in commercial container monitoring solutions when telecommunication is available---and causes frustration and confusion among users.   
Such systematic errors can occur under various situations, such as when the container contains moisture-releasing cargo, causing the sensorless predictions to consistently underestimate the relative humidity in the container.
In such cases, incorporating just a few past data points allows for correction to the predictions.

In this paper, we address these limitations.
First, we obtain 3.48 million sensor measurements from 40,677 shipments across 119 countries over a span of more than 4 years.
To the best of our knowledge, this is the largest dataset ever reported in such a study and exceeds prior works by orders of magnitude.
For each GPS coordinate, we query a local copy of the OpenStreetMaps Planet File to identify geofeatures and classify the environment around the data point.
Doing so allows us to simulate a real-world setting where the sensor readings are available on continental masses but are missing when the containers are on the seas.
Second, we adapt sensorless monitoring to use sensor data that are available after the model has been trained in conjunction with real-time exogenous variables to predict the internal conditions of the containers during information gaps.
We call this class of conditional models \emph{adaptive-sensorless} monitoring.

We test our method on 16 expanding window splits of the data.
That is, conditioned on a given month (e.g., Dec 2022), we use all the information before the month (i.e., up to and including Nov 30, 2022) as training data and use the data points in the month (i.e., Dec 1, 2022 to Dec 31, 2022) as the holdout sample.
This methodology prevents data leakage while allowing us to validate the real-world applicability of adaptive-sensorless monitoring.
Our splits span monthly from Jan 2021 to Apr 2022 (end inclusive). 
When evaluated on the holdout set of the simulated data, adaptive-sensorless models achieve average mean absolute errors (MAEs) of 2.24 $\sim$ 2.31\degC (vs 2.43\degC by sensorless) for temperature and 5.72 $\sim$ 7.09\% for relative humidity (vs 7.99\% by sensorless) and average root mean-squared errors (RMSEs) of 3.19 $\sim$ 3.26\degC for temperature (vs 3.38\degC by sensorless) and 7.70 $\sim$ 9.12\% for relative humidity (vs 10.0\% by sensorless).
These results demonstrate that adaptive-sensorless monitoring provides reliable predictions of internal conditions of shipping containers during sensor blackouts.

The remainder of this paper is organized as follows.
In \cref{sec:literature}, we review related work in container weather monitoring.
\Cref{sec:problem_statement} formalizes the problem statement.
\cref{sec:data_construction} presents our collection and processing of data whereas \cref{section:residual_correction_method} elaborates on our proposed RCM framework.
\cref{sec:experiment} details the experimental setup and the results are presented in \cref{sec:results}.
We will present brief discussions in \cref{sec:discussion} and conclude in \cref{sec:conclusion}.

\begin{figure}[t!]
  \centering
  \includegraphics[width=1.0\linewidth]{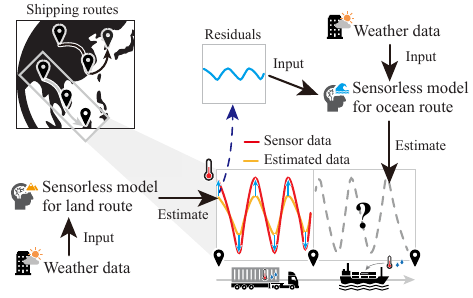}
  \caption{The concept of adaptive-sensorless monitoring for shipping containers. Adaptive-sensorless monitoring has a sensorless model for the ocean route to estimate the temperature/relative humidity inside a shipping container. The sensorless model for the ocean route uses residuals between sensor data and estimation results of the sensorless model for land route, in addition to weather data.}
  \label{fig:concept_red}
\end{figure}

\section{Literature Review}
\label{sec:literature}

Monitoring internal weather conditions within shipping containers is essential for perishable cargo. While sensor-based monitoring of temperature and relative humidity has become increasingly common, there has been limited progress in predictive modeling of these internal climate variables. Existing research has mainly focused on collecting and characterizing sensor data rather than forecasting or inference.

Several studies have reported internal conditions such as temperature and humidity along specific trade routes (\cite{singh2012measurement,scholliers2016improving,borocz2015evaluation,csavajda2019climate}), but these are typically limited in scale and scope and do not include predictive analysis. 
A notable exception is \cite{yuen2022statistical}, which modeled internal conditions using external weather data. However, their study was based on a single, static container, excluded rainy days, and did not validate on real-world shipment data.

Significant progress was made by \cite{mito2023sensorless}, which developed a sensorless monitoring model to predict internal temperature and relative humidity using only exogenous meteorological data. Leveraging what was then the largest academic dataset of its kind, comprising 85 shipments across three continents, the authors trained both linear and kernel regression models while incorporating physics-based feature engineering. Their models achieved mean absolute errors of 1.8\degC for temperature and 5.0\% for relative humidity, approaching the uncertainty range of the physical sensors themselves. However, their model is limited by the scale of the dataset, lacks geographic features, and does not account for shipment- or cargo-specific patterns in its predictions.

\section{Problem Statement}\label{sec:problem_statement}
We consider the setting where, conditioned at a point in time, we have a set of historical data $D$ that consists of live data ($D^l$) and delayed data ($D^d$): $D= \lbrace D^l, D^d \rbrace$.
`Live' means that there is no delay between the time when the data point was generated and when the data was received.
This corresponds to the case where sensor readings are transmitted to the users in real-time.
`Delayed' implies that a significant amount of time transpired before the data was obtained and corresponds to the case where the sensors lose connectivity and the data is transmitted only after connection is re-established.
These transmission properties are independent of whether the data is used for model training or inference.
When sorted temporally, the live data and delayed data can interleave.

The sensorless model developed in prior work \cite{mito2023sensorless} fits a machine learning model on all available training data (i.e., $D$).
A prediction $\hat{y}^{D}_i$ given a new data point $x_i$ is given by:
\begin{align}
    \hat{y}^{D}_i = f_0(x_i), 
\label{eqn:sensorless}
\end{align}
This works best in the case where $D= D^d$ (i.e., no live sensor data is available) and the trained model does not capture idiosyncratic features that are specific to a shipment, since the predictions are the conditional expectations over the entire data.

The objective of adaptive-sensorless monitoring---the focus of this work---is to make the best predictions for data points in $D^d$ after observing the data points in $D^l$ up to the time of prediction, which we denote as $D^l_{x_i}$:
\begin{align}
    \hat{y}^{d}_i = h(x_i, D^l_{x_i}; x_i \in D^d) 
\end{align}

The key insight is that recent sensor readings from the same shipment contain information about shipment-specific factors (cargo characteristics, container conditions, etc.) that can be used to adapt the general model for improved accuracy.
However, we face the challenge of possible domain shifts: $D^d$ and $D^l$ may have different data distributions, and models trained on $D^l$ may perform poorly when used to predict outcomes for points in $D^d$ due to geospatial differences.
We approach the problem with a three-step process termed the residual correction method (RCM), which will be elaborated on in \Cref{section:residual_correction_method} after we introduce our dataset.

\section{Data acquisition and processing}
\label{sec:data_construction}

\subsection{Sensor and meteorological measurements}
\label{sec:measurements}

\begin{figure}[!t]
  \centering
  \includegraphics[width=1.0\linewidth]{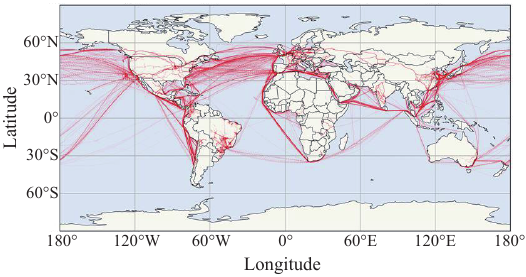}
  \caption{Geographical coverage of our sensor dataset. Redline shows the path of each shipping route. Our dataset covers almost all common shipping routes in the world.}
  \label{fig:maps_worldwide}
\end{figure}

This paper uses readings from advanced sensors with multimodal communication capabilities that are placed within 40-foot-long shipping containers, obtained from a logistics partner.
The sensors are similar to those used in Mito et al. ~\cite{mito2023sensorless} and we refer readers to that paper for details on the sensor specifications.
Our anonymized dataset consists of timed measurements of GPS coordinates, temperature and humidity inside shipping containers over journeys encompassing both land and ocean.
It contains a total of 3,487,414 data points and is, to the best of our knowledge, the largest of its kind in a published study thus far.
In contrast to the 85 shipments used in Mito et al. ~\cite{mito2023sensorless}, we consider 40,677 shipments between Jan 02, 2018 and Apr 30, 2022 across 119 countries (see \Cref{fig:maps_worldwide}).

Shipments comprise two kinds of route segments: legs and nodes.
A leg is the segment of a journey where the container is continuously transported by a mode of transportation (e.g., truck, rail, or ship), whereas a node is the segment of a journey where the container spends time between different modes of transportation (e.g., in depots or loading docks).
We train our models exclusively on data from the legs since the climate conditions in the nodes are often controlled.

\subsection{Characterizing the environment around GPS coordinates}
\label{subsection:geofeatures}

Geofeatures, also known as geographical or geological features, encode ideas or physical objects that relate to the physical space on Earth.
Geofeatures include human defined characteristics (e.g., land use such as farms or national parks), structures (e.g., buildings or roads), or geological formations (e.g., lake).

\begin{figure}[t]
  \centering
  \includegraphics[width=1.0\linewidth]{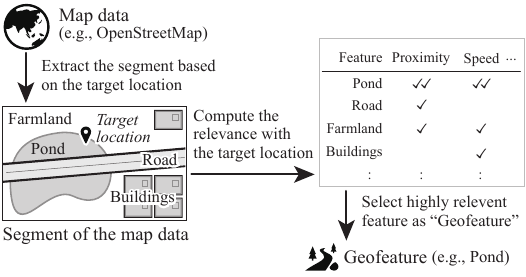}
  \caption{
  An illustrated process to create environment tags from map data.
  First, we extract an area containing the given GPS coordinate (target location) and parse the geofeatures (e.g., Farmland, Pond) around the coordinate.
  We then compute each geofeature's relevance to the GPS coordinate using a combination of heuristic rules before using them to characterize the environment around the coordinate.
  In this diagram, we give the example of a simple case where the environment is defined to be the most relevant geofeature.
  }
  \label{fig:flow_geo_features}
\end{figure}

To identify geofeatures around GPS coordinates, we create a custom library that queries a local copy of the OpenStreetMap planet file (protobuf format).
\Cref{fig:flow_geo_features} shows the steps in our pipeline.
First, the tile containing the GPS coordinates and a desired zoom level is retrieved from the database.
The tile is then decoded to obtain a set of map items, each comprising a geometric shape (e.g., Polygon, MultiPolygon, LineString) and semantic labels (e.g., ``park'', ``industrial'', ``port'', or ``residential''), among others.
For each map item, we consider the velocity vector of our GPS coordinate, the item's label, and its proximity to our GPS coordinates to compute its relevancy score.

Finally, we consider all the relevancy scores of the geofeatures to create tags for the environment surrounding the GPS coordinate.
In particular, we are interested in tagging ``roads'', ``railways'', ``urban'', ``nature'', ``port'', ``water bodies'', or ``ocean'' since they are the most relevant in the analysis of shipping routes.
Our tags are general and can be used for many applications, but their primary use in the current study is to identify data points for which the containers are traveling on land or water.

In this study, we make the simplifying assumption that the data points on land belong to $D^l$ and the data points on sea belong to $D^d$.
This is not unreasonable: cellular coverage on inhabited land allows almost instant transmission of signals, while the use of satellite communication on the sea results in delayed transmissions. 
With this assumption, our task then evolves to minimizing the differences between the predictions and the actual sensor values for data points on the ocean only.

\subsection{Data description and splits}
Recall that each sensor measurement comprises the time of the measurement, a temperature, and a relative humidity.
This is post-processed to include other meta-information such as the route identification number and the leg (or segment) identification number.

We design our train-test splits carefully in order to test the performance of the models when they are deployed under real workflows in commercial systems.
One important consideration is that while inference is computed in real-time, model training only occurs infrequently.
To emulate this, we create 16 expanding window splits of the data. 
That is, conditioned on a given month (e.g., Dec 2022), we use all the information before the month (e.g., up to and including Nov 30, 2022) as training data and use the data points in the month (i.e., Dec 1, 2022 to Dec 31, 2022) as the test data.
In this example, the cutoff date is 01 Jan 2022.
Note that in order for a data point to make it into the training set, its corresponding leg must be completed before the cutoff date.
Otherwise, it will be in the test set.
This removes the possibility of train-test contamination and also simulates how real systems work.

\begin{figure*}[!t]
  \centering
  \includegraphics[width=1.0\linewidth]{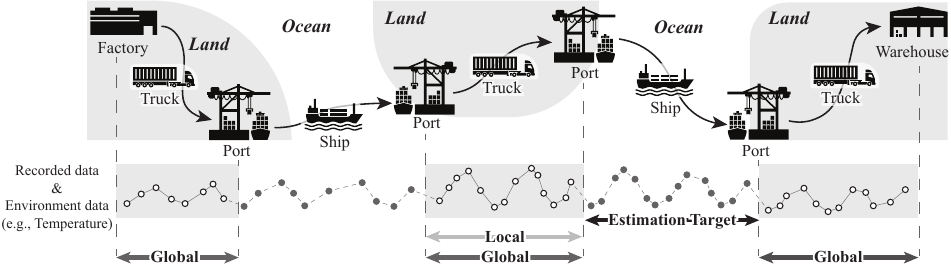}
  \caption{An illustration of different residual forecasting strategies: local versus global.
  }
  \label{fig:DefCalcShipping}
\end{figure*}

\section{Residual Correction Method}
\label{section:residual_correction_method}
This study proposes a three-step general framework that we term the residual correction method (RCM). 
First, we estimate an unconditional model $f$ using the data points $x_{s,t}^{(\ell)} \in D^l$.
Here, $s$ denotes the unique identifier of the shipment it belongs to and $t$ denotes the measurement time.
Note that $f$ differs from $f_0$ in the sensorless model in that it is trained only on $D^l$ instead of the full data $D$. 
Second, we compute the residuals $r_{s,t}^{(\ell)}$ for every $x_{s,t}^{(\ell)}$ and compute a correction factor $c_{s,t}$ with a residual weighting function $g$ (see \Cref{subsection:weighting} for further details) for each shipment $s$:
\begin{align}
  r_{s,t}^{(\ell)} &= y_{s,t}^{(\ell)} - f(x_{s,t}^{(\ell)}) && \forall x_{s,t}^{(\ell)} \in D^l\\
  c_{s,t} &= g(r_{s,1:t-1}^{(\ell)}) && \forall s
\end{align}
Note that the residuals\footnote{
A side note is that while $\mathbb{E}[r_{s,t}^{(\ell)}]=0$ for the ordinary least squares, the conditional expectation $\mathbb{E}[r_{s,t}^{(\ell)}|s]$ need not be 0.
} manifest as a univariate time series and the correction factor can be viewed as a prediction of how the shipment-specific idiosyncratic factors will evolve over time.
A simple example for $g$ is the simple average, which makes $c_{s,t}$ the time-independent bias term for the shipment.
In this case, we can simply subtract $c_{s,t}$ from $f(x_{s,t})$ to produce a prediction that is closer to the ground-truth readings.
As the third step, we estimate a separate model, $h$, using $x_{s,t}^{(o)} \in D^d$ and the $c_{s,t}$.
This three-stage modeling method allows us to adjust predictions by shipment-specific effects inferred from land data that are overlooked in the baseline model.

At inference time\footnote{Recall that we are in the setting where we only need to make predictions for $x_{s,t}^{(o)} \in D^d$.}, for any unseen data point $x_{s,t}^{(o)}$, we obtain all the observed points corresponding to the same shipment $s$ before the time $t$, compute $c_{s,t}$ using $f$ and $g$, and then compute the prediction $\hat{y}_{s,t} = h(x_{s,t}^{(o)}, c_{s,t})$.

\subsection{Model specification for $f$ and $h$}
Generally, $f$ and $h$ can be any function, including ordinary least squares (OLS) regression, kernel regression, or neural network (NN).
The main paper focuses on using OLS as it is simple to compute, fast, and interpretable.
Kernel regression is prohibitively slow to train given the quantity of data and is not considered.
We conduct experiments using simple neural networks but do not find them to be better than OLS.
Hence, they are relegated to the appendix.

\subsection{Residual weighting function, $g$}
\label{subsection:weighting}
Recall that the residuals can be represented as a time series: $r_{0}^{(\ell)}, \cdots, r_{t}^{(\ell)}$.
The subscript $s$ is dropped to simplify notation, but it should be understood that aggregation is performed at the shipment level.

Our residual weighting functions take the general form:
\begin{align*}
  g(r_{1:t}^{(\ell)})
  &= \sum_{\tau\in \{1,2,\dots,t\}}w_{\tau}r_{\tau}^{(\cdot)}, 
\end{align*}

We experiment with three weighting functions: the uniform moving average, linearly weighted average, and exponential smoothing.
The functional forms of these functions are given in \Cref{table:weighting_scheme_formula}.
As the simplest approach, the uniform moving average computes the average of past residuals. 
On the other hand, the linearly weighted average emphasizes short-term dependencies by assigning linearly increasing weights to more recent residuals.
The exponential smoothing assigns exponentially decaying weights to past residuals to favor recent residuals even more strongly.
It is worth noting that all three techniques can be implemented efficiently as streaming algorithms without the need to store the full residual history.

\begin{table}[!h]
\caption{The weighting schemes and their functional forms.}\label{table:weighting_scheme_formula}
\centering
\begin{tabular}{ll}
\toprule
Weighting functions & Functional form \\
\midrule
Uniform moving average & $w_{\tau}=1$\\
Linearly weighted average & $w_{\tau}=\frac{2}{t(t+1)} (t+1-\tau)$ \\
Exponential & $w_{\tau}=(1-\alpha)\alpha^{t-\tau}$ \\
\bottomrule
\end{tabular}
\end{table}

\subsection{Residual selection}
Each shipment may comprise a series of alternating land and sea legs, which gives us choices in \emph{which} subset of past residuals to use in the computation of $g$.
Also recall from \Cref{subsection:geofeatures} that we have made the assumption that data points on land belong to $D^l$ and the data points on sea belong to $D^d$.

We experiment with three selection schemes: local, global, and recursive.
See \cref{fig:DefCalcShipping} for an illustration of the schemes.

In the local scheme, we only include the \emph{most recent} contiguous segment of $r_{s,t}$ for $x_{s,t}^{(\ell)} \in D^l$ within the shipment. 
It best exploits the short-term spatial and temporal correlations.

In contrast, the global scheme includes \emph{all} $r_{s,t}$ for $x_{s,t}^{(\ell)} \in D^l$ within the same shipment.
Although it provides more data points and prevents overfitting to recent values, it introduces two major challenges: the sequence used for forecasting becomes unevenly spaced due to segmentation between land and ocean, and distant historical values may not be as predictive. 

To address the data scarcity of the local scheme and the segmentation of the global scheme, we consider a recursive scheme, where previously predicted corrections are fed back into the forecasting sequence.
That is, we treat prior $c_{s,t}$ as one of the inputs into $g$.\footnote{This is applicable only if there is at least one ocean segment prior to the current one.}
The scheme maintains temporal contiguity across all the segments but introduces the risk of error accumulation, especially in longer sequences.

\section{Experiment Set-up}
\label{sec:experiment}

\subsection{Measurement targets}
As previously mentioned, the task is to minimize the difference between predictions of delayed data points in $D^d$ and the ground truth (which we will eventually obtain). 
As in Mito et al. \cite{mito2023sensorless}, we focus on predicting the internal temperature and relative humidity, both of which are critical indicators of cargo conditions within shipping containers.

\subsection{Metrics}
We evaluate the performance of the models using mean absolute error (MAE) and root mean square error (RMSE) on the test set.
These metrics are chosen to capture both average prediction error and sensitivity to larger deviations.

\label{sec:model}
\subsection{Feature engineering}
In our models, we use the same list of features as Mito et al. \cite{mito2023sensorless} in addition to separating land and ocean data by the geofeatures, as discussed in \cref{subsection:geofeatures}.
This maintains comparability between this study and the sensorless model, which serves as a baseline for this study.
The list of features used is shown in \Cref{table:variables_for_models}.

Slight adaptations are made for a proxy variable named `psychrometric estimated relative humidity' when we predict relative humidity using the RCM model.
In particular, we use the predicted temperature from the conditional temperature model instead of the air temperature, which is the case in the baseline model.

\begin{table*}[!t]
\caption{
Features used in our experiments.
An asterisk (*) denotes that the feature is used in all versions of the model.
An asterisk with an `o` (i.e., *$^o$) denotes that the feature is used only on the ocean segment.
Baseline models refer to the sensorless model introduced in \cite{mito2023sensorless} and use both land and ocean data points in the training.
In contrast, conditional models train the land models only on land data points, generate the residuals, and use the residuals in training on the ocean segments.
}\label{table:variables_for_models}
\centering
\resizebox{0.9\textwidth}{!}{
    \begin{tabular}{p{4.5cm}p{9cm}cccc}
\toprule
{} & {} &Baseline&Conditional&Baseline&Conditional\\
{} & {} &temperature &temperature&humidity &humidity\\
Variable & Description &models &models &models &models\\
\midrule
\texttt{temperature} & Temperature of air at 2m above Earth's surface.& {*} & {*} & {*} & {*}\\
\texttt{solar\_radiation} & The amount of both direct and diffuse solar radiation radiation that reaches a horizontal plane at the surface of the Earth. & {*} & {*} & {*} & {*}\\
\texttt{solar\_radiation\_sq} & The square of \texttt{solar\_radiation} & {*} & {*} & {*} & {*}\\
\texttt{solar\_radiation\_sqrt} & The square root of \texttt{solar\_radiation} & {*} & {*} & {} & {}\\
\texttt{windspeed\_temperature} & $\texttt{windspeed} \cdot \texttt{temperature}$  & {*} & {*} & {*} & {*}\\
\texttt{water\_vp} & {Water vapor pressure} & {*} & {*} & {} & {}\\
\texttt{rel\_humidity} & Relative humidity of surrounding air. & {*} & {*} & {} & {}\\
\texttt{windspeed} & The neutral wind, or the mean wind speed that would be observed if there was neutral atmospheric stratification [10] & {*} & {*} & {} & {}\\
\texttt{init\_temperature} & \texttt{temperature} at the start of the leg  & {*} & {*} & {} & {}\\
\texttt{init\_rh} & \texttt{rel\_humidity} at the start of the leg& {*} & {*} & {} & {}\\
\texttt{temp\_indicator1} & $\texttt{temperature} \cdot \mathbf{1}\lbrace \frac{\texttt{temperature}}{\texttt{init\_dewpoint}} \leq \phi_1 \rbrace$  & {*} & {*} & {} & {}\\
\texttt{windspeed\_rh\_pct} & $\frac{\texttt{wind\_speed}}{\texttt{ext\_humidity}}-1$ & {*} & {*} & {} & {}\\
\texttt{windspeed\_indicator1} & $\texttt{windspeed} \cdot \mathbf{1}\lbrace \frac{\texttt{temperature}}{\texttt{init\_dewpoint}} \leq \phi_1 \rbrace$ & {*} & {*} & {} & {}\\
\texttt{temperature\_rh} & \texttt{temperature} $\cdot$ \texttt{rel\_humidity} & {} & {} & {*} & {*}\\
\texttt{rel\_dewpoint} & $\frac{\texttt{dewpoint}}{\texttt{init\_dewpoint}}-1$ & {} & {} & {*} & {*}\\
\texttt{psychro\_rh} & The relative humidity predicted by the psychrometric model given the estimated temperature and initial conditions of the container & {} & {} & {*} & {}\\
\texttt{windspeed\_indicator2} & $\texttt{windspeed} \cdot \mathbf{1}\lbrace \texttt{psychro\_rh}  \leq \phi_2 \rbrace$ & {} & {} & {*} & {*}\\
\texttt{res\_temp}&Residuals for temperature on prior land segments& {} & {*$^o$} & {} & {}\\
\texttt{res\_rh}&Residuals for humidity on prior land segements& {} & {} & {} & {*$^o$}\\
\texttt{psycho\_rh}&Psychrometric estimated relative humidity& {} & {} & {} & {*$^o$}\\

\bottomrule
\end{tabular}

}
\end{table*}

\section{Results}
\label{sec:results}

We evaluate all the variants of the RCM (either `local', `global', or `recursive' as the residual selection strategy and either `uniform', `linear', or `exponential' as the weighting strategy) against the baseline sensorless model described in \Cref{eqn:sensorless} for both temperature and relative humidity.

We also experiment with neural network models as an alternative base model in the RCM.
However, they do not show significant improvements over OLS and are hence relegated to \Cref{appendix:nn}.
More thorough experiments on the neural network architecture, regularization, and hyperparameter tuning may yield better results but are left for future work.

\subsection{Temperature}
\Cref{table:temp_mae} and \Cref{table:temp_rmse} summarize the performance of the models on temperature prediction using MAE and RMSE, respectively.
In all but two months, the best combination outperforms the baseline model.
Across all months, using linear weights on all the previous correction factors (i.e., `linear/global') yields the best results, achieving an average MAE and RMSE of 2.241 and 3.188\degC, respectively, as compared to 2.426 (MAE) and 3.376 (RMSE)\degC obtained by the base model.
These represent average improvements of 7.6\% and 5.5\% when measured using MAE and RMSE, respectively.
One thing we observed was that the ratio of RMSE to MAE for temperature prediction exceeds the typical value of 1.25\degC for a Gaussian distribution.
This indicates a heavy tail, meaning that some of the prediction errors are large (see \cref{fig:err_hist}).

\subsection{Relative Humidity}
Across all the data splits, the RCM class of models consistently outperforms the baseline model when predicting relative humidity (see \Cref{table:rh_mae} and \Cref{table:rh_rmse}).
The best-performing model (`exponential/global') achieves an average MAE of 5.72\% and an average RMSE of 7.70\%, compared to the average MAE of 7.99\% and RMSE of 10.0\% for the baseline model.
Our results are expected as the contents of the containers typically have more influence on the containers' relative humidity than temperature.
The latter is more influenced by the environment, such as the intensity of the sun's rays or the ambient environment around the containers.

\section{Discussion}
\label{sec:discussion}

The models trained using our proposed residual correction method consistently outperformed the baseline model across different configurations.
These results suggest that the RCM effectively captures cargo- or shipment-specific factors overlooked in the baseline model, as well as the temporal structure that is better leveraged through the forecasting and weighting strategies.

Whereas this study is focused on OLS and feedforward neural networks as base models, future work could explore a few different themes. First, we could explore the use of more advanced architectures such as transformers, which have shown strong performance in sequence modeling tasks. Another possible direction is to adapt the local kernel regression model from \cite{mito2023sensorless}, but the large-scale dataset used in this study presents a significant challenge for applying kernel methods, as traditional algorithms become computationally impractical at this scale.

Another promising direction is to generalize the RCM framework beyond residual correction by reinterpreting the residual as a latent variable that captures shipment-level effects. Instead of a direct error correction, the latent variable can be a multi-dimensional embedding that is learned, e.g., through neural network encoders, and then passed as auxiliary input to the ocean model. This may model more complex interactions between land and ocean. The framework proposed in this paper can be seen as an initial attempt of this more general idea. 

Beyond model design, the geospatial context is informative and can be utilized to a greater extent. For example, generating more involved geofeatures such as distance to water bodies, urban areas, or transportation routes may better explain local climates and therefore improve predictive accuracy.

\section{Conclusion}
\label{sec:conclusion}

This work introduces adaptive-sensorless monitoring as an improvement over {sensorless monitoring of shipping containers.
Our proposed residual correction method allows models to outperform their baselines on the largest dataset of its kind ever used.
This work further reinforces that monitoring of internal container conditions is possible even in the absence of real-time sensor data.
These results offer a new direction for scalable and cost-effective monitoring in global shipping.
More effective cargo monitoring, early detection of weather-related damage risks, and less dependence on in-transit connectivity should help reduce costs and minimize avoidable cargo loss.

\begin{table*}[!h]
\caption{Mean absolute error (MAE) of temperature on ocean (\degC). The residual models are labeled with ``(weighting function)/(residual selection criterion)''. Only linear models are used here.
}\label{table:temp_mae}
\resizebox{\linewidth}{!}{\begin{tabular}{|c|c|ccc|ccc|ccc|}
\hline
 & \multicolumn{1}{c|}{\textbf{Baseline}} & \multicolumn{9}{c|}{\textbf{Residual}} \\
\cline{2-11}
\textbf{YYYYMM} &  & \multicolumn{1}{c}{\textbf{uniform/local}} & \multicolumn{1}{c}{\textbf{uniform/global}} & \multicolumn{1}{c|}{\textbf{uniform/recursive}} & \multicolumn{1}{c}{\textbf{linear/local}} & \multicolumn{1}{c}{\textbf{linear/global}} & \multicolumn{1}{c|}{\textbf{linear/recursive}} & \multicolumn{1}{c}{\textbf{exp/local}} & \multicolumn{1}{c}{\textbf{exp/global}} & \multicolumn{1}{c|}{\textbf{exp/recursive}} \\
\hline
202101 & 2.569 & 2.440 & 2.387 & 2.398 & 2.442 & 2.364 & 2.393 & 2.454 & 2.389 & 2.391 \\
202102 & 2.533 & 2.429 & 2.390 & 2.411 & 2.430 & 2.365 & 2.402 & 2.440 & 2.374 & 2.377 \\
202103 & 2.495 & 2.389 & 2.359 & 2.391 & 2.390 & 2.333 & 2.376 & 2.395 & 2.330 & 2.333 \\
202104 & 2.495 & 2.357 & 2.332 & 2.377 & 2.357 & 2.305 & 2.358 & 2.357 & 2.292 & 2.295 \\
202105 & 2.473 & 2.314 & 2.286 & 2.345 & 2.312 & 2.259 & 2.323 & 2.313 & 2.247 & 2.250 \\
202106 & 2.475 & 2.310 & 2.281 & 2.338 & 2.308 & 2.253 & 2.316 & 2.301 & 2.235 & 2.237 \\
202107 & 2.482 & 2.305 & 2.271 & 2.321 & 2.303 & 2.244 & 2.302 & 2.295 & 2.228 & 2.229 \\
202108 & 2.471 & 2.300 & 2.252 & 2.281 & 2.298 & 2.225 & 2.268 & 2.290 & 2.219 & 2.221 \\
202109 & 2.467 & 2.310 & 2.242 & 2.258 & 2.309 & 2.215 & 2.249 & 2.301 & 2.222 & 2.224 \\
202110 & 2.465 & 2.336 & 2.245 & 2.264 & 2.334 & 2.215 & 2.255 & 2.325 & 2.233 & 2.235 \\
202111 & 2.477 & 2.322 & 2.206 & 2.203 & 2.317 & 2.173 & 2.194 & 2.306 & 2.211 & 2.212 \\
202112 & 2.434 & 2.275 & 2.103 & 2.057 & 2.273 & 2.081 & 2.052 & 2.265 & 2.183 & 2.183 \\
202201 & 2.321 & 2.298 & 2.183 & 2.111 & 2.299 & 2.163 & 2.112 & 2.299 & 2.248 & 2.246 \\
202202 & 2.250 & 2.280 & 2.342 & 2.289 & 2.285 & 2.321 & 2.282 & 2.329 & 2.345 & 2.343 \\
202203 & 2.165 & 2.187 & 2.266 & 2.247 & 2.188 & 2.249 & 2.239 & 2.250 & 2.271 & 2.270 \\
202204 & 2.250 & 2.112 & 2.097 & 2.085 & 2.112 & 2.087 & 2.076 & 2.157 & 2.180 & 2.183 \\
\hline
Average & 2.426 & 2.310 & 2.265 & 2.274 & 2.310 & 2.241 & 2.262 & 2.317 & 2.263 & 2.264 \\
\hline
\end{tabular}}
\end{table*}

\begin{table*}[!h]
\caption{Root mean square error (RMSE) of temperature on ocean (\degC). The residual models are labeled with ``(weighting function)/(residual selection criterion)''. Only linear models are used here.}\label{table:temp_rmse}
\resizebox{\linewidth}{!}{\begin{tabular}{|c|c|ccc|ccc|ccc|}
\hline
 & \multicolumn{1}{c|}{\textbf{Baseline}} & \multicolumn{9}{c|}{\textbf{Residual}} \\
\cline{2-11}
\textbf{YYYYMM} &  & \multicolumn{1}{c}{\textbf{uniform/local}} & \multicolumn{1}{c}{\textbf{uniform/global}} & \multicolumn{1}{c|}{\textbf{uniform/recursive}} & \multicolumn{1}{c}{\textbf{linear/local}} & \multicolumn{1}{c}{\textbf{linear/global}} & \multicolumn{1}{c|}{\textbf{linear/recursive}} & \multicolumn{1}{c}{\textbf{exp/local}} & \multicolumn{1}{c}{\textbf{exp/global}} & \multicolumn{1}{c|}{\textbf{exp/recursive}} \\
\hline
202101 & 3.609 & 3.445 & 3.396 & 3.444 & 3.446 & 3.373 & 3.416 & 3.462 & 3.391 & 3.393 \\
202102 & 3.567 & 3.427 & 3.389 & 3.444 & 3.428 & 3.364 & 3.413 & 3.441 & 3.370 & 3.372 \\
202103 & 3.523 & 3.368 & 3.333 & 3.400 & 3.368 & 3.309 & 3.364 & 3.378 & 3.306 & 3.308 \\
202104 & 3.516 & 3.327 & 3.296 & 3.378 & 3.327 & 3.272 & 3.337 & 3.333 & 3.261 & 3.263 \\
202105 & 3.500 & 3.282 & 3.248 & 3.345 & 3.281 & 3.224 & 3.301 & 3.287 & 3.214 & 3.216 \\
202106 & 3.502 & 3.276 & 3.240 & 3.335 & 3.274 & 3.216 & 3.291 & 3.273 & 3.200 & 3.202 \\
202107 & 3.502 & 3.256 & 3.217 & 3.306 & 3.255 & 3.193 & 3.264 & 3.253 & 3.179 & 3.179 \\
202108 & 3.486 & 3.241 & 3.190 & 3.260 & 3.240 & 3.166 & 3.224 & 3.238 & 3.160 & 3.161 \\
202109 & 3.424 & 3.231 & 3.157 & 3.204 & 3.230 & 3.133 & 3.178 & 3.224 & 3.135 & 3.136 \\
202110 & 3.292 & 3.229 & 3.124 & 3.161 & 3.227 & 3.100 & 3.148 & 3.212 & 3.106 & 3.109 \\
202111 & 3.298 & 3.215 & 3.087 & 3.100 & 3.209 & 3.063 & 3.091 & 3.190 & 3.083 & 3.084 \\
202112 & 3.257 & 3.144 & 2.998 & 2.973 & 3.142 & 2.979 & 2.967 & 3.133 & 3.048 & 3.048 \\
202201 & 3.162 & 3.183 & 3.123 & 3.071 & 3.185 & 3.104 & 3.074 & 3.189 & 3.145 & 3.145 \\
202202 & 3.131 & 3.219 & 3.306 & 3.289 & 3.222 & 3.287 & 3.281 & 3.257 & 3.272 & 3.270 \\
202203 & 3.085 & 3.113 & 3.222 & 3.241 & 3.112 & 3.208 & 3.234 & 3.163 & 3.184 & 3.185 \\
202204 & 3.161 & 3.022 & 3.032 & 3.050 & 3.023 & 3.022 & 3.036 & 3.067 & 3.087 & 3.090 \\
\hline
Average & 3.376 & 3.249 & 3.210 & 3.250 & 3.248 & 3.188 & 3.226 & 3.256 & 3.196 & 3.198 \\
\hline
\end{tabular}}
\end{table*}

\begin{table*}[!h]
\caption{Mean absolute error (MAE) of relative humidity on ocean (\%). The residual models are labeled with ``(weighting function)/(residual selection criterion)''. Only linear models are used here.}\label{table:rh_mae}
\resizebox{\linewidth}{!}{\begin{tabular}{|c|c|ccc|ccc|ccc|}
\hline
 & \multicolumn{1}{c|}{\textbf{Baseline}} & \multicolumn{9}{c|}{\textbf{Residual}} \\
\cline{2-11}
\textbf{YYYYMM} &  & \multicolumn{1}{c}{\textbf{uniform/local}} & \multicolumn{1}{c}{\textbf{uniform/global}} & \multicolumn{1}{c|}{\textbf{uniform/recursive}} & \multicolumn{1}{c}{\textbf{linear/local}} & \multicolumn{1}{c}{\textbf{linear/global}} & \multicolumn{1}{c|}{\textbf{linear/recursive}} & \multicolumn{1}{c}{\textbf{exp/local}} & \multicolumn{1}{c}{\textbf{exp/global}} & \multicolumn{1}{c|}{\textbf{exp/recursive}} \\
\hline
202101 & 8.503 & 6.152 & 6.141 & 7.416 & 6.113 & 5.996 & 6.676 & 6.151 & 5.975 & 6.001 \\
202102 & 8.591 & 6.102 & 6.058 & 7.325 & 6.063 & 5.914 & 6.597 & 6.087 & 5.912 & 5.938 \\
202103 & 8.686 & 5.994 & 5.938 & 7.276 & 5.955 & 5.790 & 6.507 & 5.978 & 5.788 & 5.814 \\
202104 & 8.766 & 5.910 & 5.861 & 7.163 & 5.870 & 5.717 & 6.411 & 5.892 & 5.706 & 5.728 \\
202105 & 8.783 & 5.836 & 5.808 & 7.176 & 5.794 & 5.660 & 6.396 & 5.813 & 5.632 & 5.656 \\
202106 & 8.641 & 5.741 & 5.743 & 6.991 & 5.694 & 5.578 & 6.267 & 5.711 & 5.532 & 5.558 \\
202107 & 8.378 & 5.703 & 5.695 & 6.971 & 5.658 & 5.531 & 6.212 & 5.679 & 5.489 & 5.512 \\
202108 & 8.199 & 5.620 & 5.602 & 6.583 & 5.574 & 5.439 & 5.926 & 5.586 & 5.386 & 5.405 \\
202109 & 8.094 & 5.595 & 5.574 & 6.583 & 5.546 & 5.402 & 5.894 & 5.554 & 5.344 & 5.365 \\
202110 & 7.977 & 5.526 & 5.495 & 6.238 & 5.473 & 5.318 & 5.670 & 5.479 & 5.256 & 5.276 \\
202111 & 7.536 & 5.644 & 5.542 & 6.091 & 5.584 & 5.362 & 5.626 & 5.582 & 5.327 & 5.349 \\
202112 & 7.165 & 5.926 & 5.850 & 6.214 & 5.858 & 5.625 & 5.810 & 5.847 & 5.586 & 5.610 \\
202201 & 6.619 & 6.071 & 6.601 & 7.283 & 6.000 & 6.313 & 6.709 & 5.984 & 5.830 & 5.863 \\
202202 & 7.381 & 6.514 & 7.896 & 9.204 & 6.449 & 7.475 & 8.443 & 6.488 & 6.452 & 6.504 \\
202203 & 7.314 & 6.419 & 7.323 & 8.308 & 6.361 & 6.986 & 7.809 & 6.404 & 6.347 & 6.395 \\
202204 & 7.276 & 6.046 & 6.227 & 6.654 & 5.999 & 5.959 & 6.248 & 6.121 & 5.940 & 5.957 \\
\hline
Average & 7.994 & 5.925 & 6.085 & 7.092 & 5.874 & 5.879 & 6.450 & 5.897 & 5.719 & 5.746 \\
\hline
\end{tabular}}
\end{table*}

\begin{table*}[!h]
\caption{Root mean square error (RMSE) of relative humidity on ocean (\%). The residual models are labeled with ``(weighting function)/(residual selection criterion)''. Only linear models are used here.}\label{table:rh_rmse}
\resizebox{\linewidth}{!}{\begin{tabular}{|c|c|ccc|ccc|ccc|}
\hline
 & \multicolumn{1}{c|}{\textbf{Baseline}} & \multicolumn{9}{c|}{\textbf{Residual}} \\
\cline{2-11}
\textbf{YYYYMM} &  & \multicolumn{1}{c}{\textbf{uniform/local}} & \multicolumn{1}{c}{\textbf{uniform/global}} & \multicolumn{1}{c|}{\textbf{uniform/recursive}} & \multicolumn{1}{c}{\textbf{linear/local}} & \multicolumn{1}{c}{\textbf{linear/global}} & \multicolumn{1}{c|}{\textbf{linear/recursive}} & \multicolumn{1}{c}{\textbf{exp/local}} & \multicolumn{1}{c}{\textbf{exp/global}} & \multicolumn{1}{c|}{\textbf{exp/recursive}} \\
\hline
202101 & 10.554 & 8.209 & 8.290 & 9.532 & 8.168 & 8.132 & 8.790 & 8.194 & 8.055 & 8.090 \\
202102 & 10.653 & 8.150 & 8.194 & 9.418 & 8.107 & 8.038 & 8.692 & 8.117 & 7.981 & 8.016 \\
202103 & 10.762 & 8.039 & 8.058 & 9.324 & 7.996 & 7.901 & 8.561 & 8.005 & 7.850 & 7.883 \\
202104 & 10.833 & 7.964 & 7.986 & 9.212 & 7.922 & 7.828 & 8.468 & 7.931 & 7.776 & 7.806 \\
202105 & 10.828 & 7.854 & 7.898 & 9.186 & 7.807 & 7.733 & 8.410 & 7.812 & 7.659 & 7.690 \\
202106 & 10.674 & 7.743 & 7.819 & 8.996 & 7.691 & 7.636 & 8.268 & 7.692 & 7.540 & 7.574 \\
202107 & 10.358 & 7.679 & 7.750 & 8.949 & 7.629 & 7.565 & 8.190 & 7.632 & 7.466 & 7.498 \\
202108 & 10.162 & 7.540 & 7.618 & 8.586 & 7.488 & 7.426 & 7.913 & 7.482 & 7.307 & 7.336 \\
202109 & 10.051 & 7.510 & 7.591 & 8.579 & 7.454 & 7.388 & 7.879 & 7.444 & 7.258 & 7.290 \\
202110 & 9.926 & 7.413 & 7.488 & 8.245 & 7.353 & 7.278 & 7.647 & 7.336 & 7.137 & 7.166 \\
202111 & 9.536 & 7.510 & 7.541 & 8.106 & 7.443 & 7.322 & 7.601 & 7.417 & 7.192 & 7.223 \\
202112 & 9.290 & 7.765 & 7.861 & 8.225 & 7.694 & 7.612 & 7.789 & 7.674 & 7.469 & 7.499 \\
202201 & 8.656 & 7.990 & 8.702 & 9.426 & 7.915 & 8.368 & 8.798 & 7.887 & 7.782 & 7.821 \\
202202 & 9.335 & 8.513 & 9.870 & 11.109 & 8.439 & 9.411 & 10.325 & 8.456 & 8.438 & 8.493 \\
202203 & 9.233 & 8.419 & 9.352 & 10.267 & 8.351 & 8.978 & 9.747 & 8.375 & 8.339 & 8.388 \\
202204 & 9.296 & 8.101 & 8.354 & 8.790 & 8.040 & 8.067 & 8.382 & 8.153 & 7.995 & 8.017 \\
\hline
Average & 10.01 & 7.900 & 8.148 & 9.122 & 7.844 & 7.918 & 8.466 & 7.850 & 7.703 & 7.737 \\
\hline
\end{tabular}}
\end{table*}

\newpage
\bibliographystyle{IEEEtran}
\bibliography{reference.bib}

@article{viitanen2007improved,
  title={Improved model to predict mold growth in building materials},
  author={Viitanen, Hannu and Ojanen, Tuomo},
  journal={Thermal Performance of the Exterior Envelopes of Whole Buildings X--Proceedings CD},
  pages={2--7},
  year={2007}
}

@article{borocz2015evaluation,
  title={Evaluation of distribution environment in LTL shipment between central Europe and South Africa},
  author={Borocz, Peter and Singh, Paul and Singh, Jay},
  journal={Journal of Applied Packaging Research},
  volume={7},
  number={2},
  pages={3},
  year={2015}
}

@article{csavajda2019climate,
  title={Climate conditions in ISO container shipments from Hungary to South Africa and Asia},
  author={Csavajda, P{\'e}ter and B{\"o}r{\"o}cz, P{\'e}ter},
  journal={Periodica Polytechnica Transportation Engineering},
  volume={47},
  number={3},
  pages={233--241},
  year={2019}
}

@article{scholliers2016improving,
  title={Improving the security of containers in port related supply chains},
  author={Scholliers, Johan and Permala, Antti and Toivonen, Sirra and Salmela, Hannu},
  journal={Transportation research procedia},
  volume={14},
  pages={1374--1383},
  year={2016},
  publisher={Elsevier}
}

@article{unctd2022review,
    title={Review of maritime transport 2022},
    author={{United Nations Conference on Trade and Development}},
    year={2022},
    publisher={United Nations Publications},
    note={{Available:} \href{https://unctad.org/system/files/official-document/rmt2022_en.pdf}{https://unctad.org{\slash}system{\slash}files{\slash}official-document{\slash}rmt2022{\slash}{\_}en.pdf},\space Accessed on Aug 8, 2023},
}

@article{viitanen2015mold,
  title={Mold risk classification based on comparative evaluation of two established growth models},
  author={Viitanen, Hannu and Krus, M and Ojanen, Tuomo and Eitner, V and Zirkelbach, Daniel},
  journal={Energy Procedia},
  volume={78},
  pages={1425--1430},
  year={2015},
  publisher={Elsevier}
}

@article{singh2012measurement,
  title={Measurement and analysis of vibration and temperature levels in global intermodal container shipments on truck, rail and ship},
  author={Singh, S Paul and Saha, K and Singh, J and Sandhu, APS},
  journal={Packaging Technology and Science},
  volume={25},
  number={3},
  pages={149--160},
  year={2012},
  publisher={Wiley Online Library}
}

@inproceedings{mito2023sensorless,
  title={Sensorless Monitoring of Shipping Containers},
  author={Mito, Misaki and Wong, Chi Heem and Chakrabarti, Arnab},
  booktitle={2023 IEEE International Conference on Big Data (BigData)},
  pages={1782--1792},
  year={2023},
  organization={IEEE}
}

@article{yuen2022statistical,
  title={Statistical estimation of container condensation in marine transportation between Far East Asia and Europe},
  author={Yuen, Ping Chi and Sasa, Kenji and Kawahara, Hideo and Chen, Chen},
  journal={The Journal of Navigation},
  volume={75},
  number={1},
  pages={176--199},
  year={2022},
  publisher={Cambridge University Press}
}

\clearpage
\onecolumn
\section*{APPENDIX}
\setcounter{table}{0}
\setcounter{figure}{0}
\setcounter{section}{0}
\renewcommand\thefigure{A\arabic{figure}}
\renewcommand\thetable{A\arabic{table}}
\renewcommand\thesection{A\arabic{section}}

\section{Neural network model}\label{appendix:nn}
In addition to linear models, we experiment with neural networks as specifications of $f$ and $g$.
The neural networks ingest the exact same features as the linear models.

We implement small fully connected networks ($<$250k parameters) that have 2 output heads to predict the temperature and relative humidity concurrently.
The model consists of a temperature subnetwork and a relative humidity subnetwork, both of which take the input feature vector and pass it through a sequence of residual fully connected layers.
More parameters are allocated to the relative humidity subnetwork as we have determined that it is more difficult to learn.
The psychrometric feature is computed from the output of the temperature subnetwork and concatenated as an additional input to the relative humidity subnetwork.
The psychrometry library (non-learnable) is reimplemented in PyTorch so that backpropagation passes through seamlessly.
In addition, the relative humidity output is clamped between 0 and 100 to maintain feasibility.
Residual connections are used through the entire neural network and a dropout rate of 5\% is applied to the penultimate layers before the prediction heads.
The network was trained using the Adam optimizer with a learning rate of 0.001, and we use mean squared error as the loss function. 
Each training was run for 200 epochs.

The results are presented in \Cref{table:nn_temp_mae,table:nn_temp_rmse,table:nn_rh_mae,table:nn_rh_rmse}.
When comparing against the linear models, we see that the results are mixed; neural networks perform better in some cases but only marginally. 
As such, for simplicity, we decide to only report linear models in the main paper.
Nonetheless, improvements of the RCM against the baseline model were observed, which validates the RCM framework.

It is well known that deep neural networks show performance gains as more and more data is added, which is a major reason for their success with web-scale data.
Our dataset, although large, is not even close to web-scale. With a combination of the right architecture and sufficient regularization, neural network models should produce results comparable to or better than OLS models. This is left for future work.

\begin{table*}[!h]
\caption{Mean absolute error (MAE) of temperature on ocean for neural networks (\degC). The residual models are labeled with ``(weighting function)/(residual selection criterion)''.}\label{table:nn_temp_mae}
\resizebox{\linewidth}{!}{\begin{tabular}{|l|cc|ccc|ccc|ccc|}
\hline
 & \multicolumn{2}{c|}{\textbf{Baseline}} & \multicolumn{9}{c|}{\textbf{Residual}} \\
\cline{2-12}
 & \multicolumn{1}{c}{\textbf{Linear}} & \multicolumn{1}{c|}{\textbf{NN}} & \multicolumn{9}{c|}{\textbf{NN}} \\
\cline{4-12}
\textbf{Date} &  &  & \multicolumn{1}{c}{\textbf{uniform/local}} & \multicolumn{1}{c}{\textbf{uniform/global}} & \multicolumn{1}{c|}{\textbf{uniform/recursive}} & \multicolumn{1}{c}{\textbf{linear/local}} & \multicolumn{1}{c}{\textbf{linear/global}} & \multicolumn{1}{c|}{\textbf{linear/recursive}} & \multicolumn{1}{c}{\textbf{exp/local}} & \multicolumn{1}{c}{\textbf{exp/global}} & \multicolumn{1}{c|}{\textbf{exp/recursive}} \\
\hline
202101 & 2.569 & 2.328 & 2.071 & 2.055 & 2.095 & 2.107 & 2.111 & 2.110 & 2.123 & 2.064 & 1.984 \\
202102 & 2.533 & 3.227 & 2.156 & 2.046 & 2.006 & 2.092 & 2.097 & 2.080 & 2.084 & 1.991 & 2.020 \\
202103 & 2.495 & 2.393 & 2.133 & 2.145 & 1.963 & 2.091 & 2.079 & 2.002 & 2.228 & 2.023 & 2.028 \\
202104 & 2.495 & 2.419 & 1.988 & 1.946 & 2.125 & 1.971 & 1.999 & 1.971 & 1.952 & 1.930 & 1.886 \\
202105 & 2.473 & 2.445 & 1.951 & 1.983 & 1.921 & 1.958 & 1.914 & 2.005 & 1.940 & 1.932 & 1.977 \\
202106 & 2.475 & 2.472 & 1.958 & 1.920 & 1.976 & 1.901 & 1.907 & 1.961 & 1.926 & 1.888 & 1.853 \\
202107 & 2.482 & 2.240 & 1.853 & 1.845 & 2.001 & 1.882 & 1.873 & 1.933 & 1.919 & 1.841 & 1.881 \\
202108 & 2.471 & 2.582 & 1.924 & 1.935 & 2.176 & 1.938 & 1.951 & 1.942 & 1.958 & 1.927 & 1.867 \\
202109 & 2.467 & 2.352 & 1.991 & 1.912 & 2.118 & 1.983 & 1.917 & 2.075 & 1.978 & 1.941 & 1.927 \\
202110 & 2.465 & 2.770 & 2.045 & 2.036 & 1.907 & 2.040 & 1.898 & 2.069 & 1.966 & 2.036 & 1.927 \\
202111 & 2.477 & 2.319 & 1.987 & 1.893 & 1.837 & 1.974 & 1.916 & 1.839 & 1.975 & 1.910 & 1.897 \\
202112 & 2.434 & 2.265 & 1.876 & 1.815 & 1.912 & 1.874 & 1.784 & 1.813 & 1.861 & 1.846 & 1.896 \\
202201 & 2.321 & 2.625 & 2.247 & 2.032 & 1.897 & 2.004 & 2.151 & 1.968 & 2.067 & 2.151 & 2.013 \\
202202 & 2.250 & 3.052 & 3.262 & 3.168 & 2.968 & 3.135 & 3.220 & 2.608 & 3.002 & 2.891 & 3.161 \\
202203 & 2.165 & 2.812 & 2.929 & 2.780 & 2.800 & 2.846 & 2.890 & 2.763 & 2.670 & 2.808 & 2.784 \\
202204 & 2.250 & 2.761 & 2.571 & 2.719 & 2.502 & 2.614 & 2.765 & 2.408 & 2.546 & 2.611 & 2.667 \\
\hline
Average & 2.426 & 2.566 & 2.184 & 2.139 & 2.138 & 2.151 & 2.155 & 2.097 & 2.137 & 2.112 & 2.111 \\
\hline
\end{tabular}}
\end{table*}

\begin{table*}[!h]
\caption{Root mean squared error (RMSE) of temperature on ocean for neural networks (\degC). The residual models are labeled with ``(weighting function)/(residual selection criterion)''.
}\label{table:nn_temp_rmse}
\resizebox{\linewidth}{!}{\begin{tabular}{|l|cc|ccc|ccc|ccc|}
\hline
 & \multicolumn{2}{c|}{\textbf{Baseline}} & \multicolumn{9}{c|}{\textbf{Residual}} \\
\cline{2-12}
 & \multicolumn{1}{c}{\textbf{Linear}} & \multicolumn{1}{c|}{\textbf{NN}} & \multicolumn{9}{c|}{\textbf{NN}} \\
\cline{4-12}
\textbf{Date} &  &  & \multicolumn{1}{c}{\textbf{uniform/local}} & \multicolumn{1}{c}{\textbf{uniform/global}} & \multicolumn{1}{c|}{\textbf{uniform/recursive}} & \multicolumn{1}{c}{\textbf{linear/local}} & \multicolumn{1}{c}{\textbf{linear/global}} & \multicolumn{1}{c|}{\textbf{linear/recursive}} & \multicolumn{1}{c}{\textbf{exp/local}} & \multicolumn{1}{c}{\textbf{exp/global}} & \multicolumn{1}{c|}{\textbf{exp/recursive}} \\
\hline
202101 & 3.609 & 3.496 & 3.127 & 3.095 & 3.205 & 3.161 & 3.166 & 3.192 & 3.165 & 3.089 & 3.052 \\
202102 & 3.567 & 4.209 & 3.162 & 3.066 & 3.104 & 3.165 & 3.080 & 3.161 & 3.138 & 3.013 & 3.050 \\
202103 & 3.523 & 3.568 & 3.184 & 3.083 & 3.041 & 3.131 & 3.096 & 3.055 & 3.199 & 3.034 & 3.056 \\
202104 & 3.516 & 3.599 & 3.009 & 2.965 & 3.248 & 2.937 & 2.974 & 3.011 & 2.942 & 2.899 & 2.887 \\
202105 & 3.500 & 3.631 & 2.955 & 2.958 & 2.980 & 2.959 & 2.883 & 3.036 & 2.939 & 2.920 & 2.978 \\
202106 & 3.502 & 3.622 & 2.995 & 2.907 & 3.054 & 2.928 & 2.893 & 3.007 & 2.958 & 2.887 & 2.859 \\
202107 & 3.502 & 3.411 & 2.872 & 2.820 & 2.990 & 2.885 & 2.855 & 2.928 & 2.885 & 2.838 & 2.843 \\
202108 & 3.486 & 3.721 & 2.945 & 2.909 & 3.092 & 2.933 & 2.919 & 2.950 & 2.957 & 2.901 & 2.856 \\
202109 & 3.424 & 3.470 & 2.970 & 2.848 & 3.062 & 2.966 & 2.865 & 3.018 & 2.955 & 2.898 & 2.884 \\
202110 & 3.292 & 3.721 & 3.004 & 2.947 & 2.881 & 3.006 & 2.826 & 2.976 & 2.913 & 2.901 & 2.855 \\
202111 & 3.298 & 3.312 & 2.953 & 2.823 & 2.837 & 2.947 & 2.860 & 2.812 & 2.936 & 2.851 & 2.836 \\
202112 & 3.257 & 3.241 & 2.862 & 2.819 & 2.861 & 2.865 & 2.776 & 2.781 & 2.848 & 2.812 & 2.922 \\
202201 & 3.162 & 3.609 & 3.329 & 3.145 & 2.988 & 3.089 & 3.248 & 3.066 & 3.164 & 3.218 & 3.098 \\
202202 & 3.131 & 4.130 & 4.432 & 4.430 & 4.064 & 4.326 & 4.540 & 3.754 & 4.200 & 4.012 & 4.398 \\
202203 & 3.085 & 3.773 & 4.057 & 4.015 & 3.904 & 3.938 & 4.055 & 3.948 & 3.696 & 3.889 & 3.876 \\
202204 & 3.161 & 3.763 & 3.651 & 3.785 & 3.536 & 3.674 & 3.884 & 3.445 & 3.570 & 3.591 & 3.657 \\
\hline
Average & 3.376 & 3.642 & 3.219 & 3.163 & 3.178 & 3.182 & 3.182 & 3.134 & 3.154 & 3.110 & 3.132 \\
\hline
\end{tabular}}
\end{table*}

\begin{table*}[!h]
\caption{Mean absolute error (MAE) of relative humidity on ocean for neural networks (\%). The residual models are labeled with ``(weighting function)/(residual selection criterion)''.}\label{table:nn_rh_mae}
\resizebox{\linewidth}{!}{\begin{tabular}{|l|cc|ccc|ccc|ccc|}
\hline
 & \multicolumn{2}{c|}{\textbf{Baseline}} & \multicolumn{9}{c|}{\textbf{Residual}} \\
\cline{2-12}
 & \multicolumn{1}{c}{\textbf{Linear}} & \multicolumn{1}{c|}{\textbf{NN}} & \multicolumn{9}{c|}{\textbf{NN}} \\
\cline{4-12}
\textbf{Date} &  &  & \multicolumn{1}{c}{\textbf{uniform/local}} & \multicolumn{1}{c}{\textbf{uniform/global}} & \multicolumn{1}{c|}{\textbf{uniform/recursive}} & \multicolumn{1}{c}{\textbf{linear/local}} & \multicolumn{1}{c}{\textbf{linear/global}} & \multicolumn{1}{c|}{\textbf{linear/recursive}} & \multicolumn{1}{c}{\textbf{exp/local}} & \multicolumn{1}{c}{\textbf{exp/global}} & \multicolumn{1}{c|}{\textbf{exp/recursive}} \\
\hline
202101 & 8.503 & 8.588 & 5.754 & 5.599 & 6.518 & 5.647 & 5.498 & 6.143 & 5.709 & 5.646 & 5.653 \\
202102 & 8.591 & 8.728 & 5.746 & 5.589 & 6.593 & 5.659 & 5.536 & 6.121 & 5.669 & 5.667 & 5.569 \\
202103 & 8.686 & 8.161 & 5.644 & 5.512 & 6.605 & 5.651 & 5.424 & 6.098 & 5.759 & 5.623 & 5.498 \\
202104 & 8.766 & 8.580 & 5.523 & 5.566 & 6.364 & 5.514 & 5.328 & 6.366 & 5.461 & 5.454 & 5.430 \\
202105 & 8.783 & 8.484 & 5.716 & 5.521 & 6.445 & 5.694 & 5.585 & 6.240 & 5.686 & 5.587 & 5.602 \\
202106 & 8.641 & 8.526 & 5.726 & 5.526 & 6.757 & 5.690 & 5.420 & 6.337 & 5.746 & 5.474 & 5.518 \\
202107 & 8.378 & 8.498 & 5.471 & 5.462 & 6.705 & 5.491 & 5.245 & 6.438 & 5.564 & 5.403 & 5.425 \\
202108 & 8.199 & 8.279 & 5.450 & 5.364 & 6.996 & 5.444 & 5.199 & 6.405 & 5.445 & 5.223 & 5.272 \\
202109 & 8.094 & 7.469 & 5.390 & 5.338 & 7.293 & 5.314 & 5.129 & 6.525 & 5.372 & 5.210 & 5.209 \\
202110 & 7.977 & 7.883 & 5.470 & 5.230 & 6.489 & 5.434 & 5.337 & 6.348 & 5.404 & 5.306 & 5.178 \\
202111 & 7.536 & 7.094 & 5.180 & 4.893 & 6.200 & 5.079 & 4.755 & 5.258 & 5.007 & 4.856 & 4.847 \\
202112 & 7.165 & 7.060 & 5.231 & 5.162 & 6.298 & 5.160 & 5.033 & 5.666 & 5.190 & 5.090 & 5.424 \\
202201 & 6.619 & 6.833 & 5.413 & 5.358 & 6.389 & 5.393 & 5.216 & 5.943 & 5.444 & 5.326 & 5.490 \\
202202 & 7.381 & 7.985 & 6.388 & 6.184 & 6.817 & 6.283 & 6.032 & 6.277 & 6.284 & 6.139 & 6.197 \\
202203 & 7.314 & 7.167 & 6.234 & 5.950 & 6.335 & 6.201 & 5.931 & 6.109 & 6.225 & 6.062 & 6.163 \\
202204 & 7.276 & 7.146 & 6.006 & 6.150 & 6.207 & 6.032 & 5.954 & 6.107 & 6.071 & 5.994 & 5.960 \\
\hline
Average & 7.994 & 7.905 & 5.646 & 5.525 & 6.563 & 5.605 & 5.414 & 6.149 & 5.627 & 5.504 & 5.527\\
\hline
\end{tabular}}
\end{table*}

\begin{table*}[!h]
\caption{Root mean squared error (RMSE) of relative humidity on ocean for neural networks (\%). The residual models are labeled with ``(weighting function)/(residual selection criterion)''.}\label{table:nn_rh_rmse}
\resizebox{\linewidth}{!}{\begin{tabular}{|l|cc|ccc|ccc|ccc|}
\hline
 & \multicolumn{2}{c|}{\textbf{Baseline}} & \multicolumn{9}{c|}{\textbf{Residual}} \\
\cline{2-12}
 & \multicolumn{1}{c}{\textbf{Linear}} & \multicolumn{1}{c|}{\textbf{NN}} & \multicolumn{9}{c|}{\textbf{NN}} \\
\cline{4-12}
\textbf{Date} &  &  & \multicolumn{1}{c}{\textbf{uniform/local}} & \multicolumn{1}{c}{\textbf{uniform/global}} & \multicolumn{1}{c|}{\textbf{uniform/recursive}} & \multicolumn{1}{c}{\textbf{linear/local}} & \multicolumn{1}{c}{\textbf{linear/global}} & \multicolumn{1}{c|}{\textbf{linear/recursive}} & \multicolumn{1}{c}{\textbf{exp/local}} & \multicolumn{1}{c}{\textbf{exp/global}} & \multicolumn{1}{c|}{\textbf{exp/recursive}} \\
\hline
202101 & 10.554 & 10.838 & 7.768 & 7.634 & 8.761 & 7.665 & 7.517 & 8.227 & 7.731 & 7.687 & 7.751 \\
202102 & 10.653 & 10.950 & 7.782 & 7.616 & 8.773 & 7.642 & 7.601 & 8.147 & 7.682 & 7.707 & 7.602 \\
202103 & 10.762 & 10.339 & 7.590 & 7.542 & 8.661 & 7.655 & 7.419 & 8.070 & 7.788 & 7.647 & 7.490 \\
202104 & 10.833 & 10.865 & 7.502 & 7.691 & 8.430 & 7.552 & 7.378 & 8.429 & 7.456 & 7.508 & 7.491 \\
202105 & 10.828 & 10.773 & 7.668 & 7.549 & 8.542 & 7.647 & 7.618 & 8.222 & 7.655 & 7.552 & 7.575 \\
202106 & 10.674 & 10.738 & 7.676 & 7.549 & 8.792 & 7.642 & 7.444 & 8.372 & 7.697 & 7.426 & 7.494 \\
202107 & 10.358 & 10.710 & 7.377 & 7.453 & 8.834 & 7.394 & 7.217 & 8.458 & 7.500 & 7.318 & 7.350 \\
202108 & 10.162 & 10.476 & 7.277 & 7.292 & 9.035 & 7.291 & 7.113 & 8.404 & 7.297 & 7.090 & 7.147 \\
202109 & 10.051 & 9.595 & 7.250 & 7.275 & 9.379 & 7.187 & 7.043 & 8.533 & 7.235 & 7.072 & 7.079 \\
202110 & 9.926 & 10.063 & 7.288 & 7.129 & 8.663 & 7.274 & 7.310 & 8.669 & 7.284 & 7.180 & 7.044 \\
202111 & 9.536 & 9.184 & 6.979 & 6.757 & 8.376 & 6.884 & 6.596 & 7.167 & 6.792 & 6.641 & 6.663 \\
202112 & 9.290 & 9.138 & 7.160 & 7.134 & 8.492 & 7.090 & 7.001 & 7.751 & 7.131 & 7.016 & 7.243 \\
202201 & 8.656 & 9.018 & 7.341 & 7.322 & 8.612 & 7.352 & 7.160 & 8.071 & 7.390 & 7.250 & 7.484 \\
202202 & 9.335 & 10.229 & 8.437 & 8.243 & 8.876 & 8.281 & 8.084 & 8.292 & 8.291 & 8.110 & 8.189 \\
202203 & 9.233 & 9.249 & 8.324 & 8.010 & 8.451 & 8.285 & 8.007 & 8.190 & 8.290 & 8.119 & 8.249 \\
202204 & 9.296 & 9.146 & 8.140 & 8.270 & 8.321 & 8.172 & 8.064 & 8.255 & 8.190 & 8.153 & 8.131 \\
\hline
Average & 10.01 & 10.08 & 7.597 & 7.529 & 8.687 & 7.563 & 7.411 & 8.203 & 7.588 & 7.467 & 7.499 \\
\hline
\end{tabular}}
\end{table*}

\newpage
\section{Distribution of temperature prediction error}
\begin{figure}[!h]
  \centering
  \includegraphics[width=0.5\linewidth]{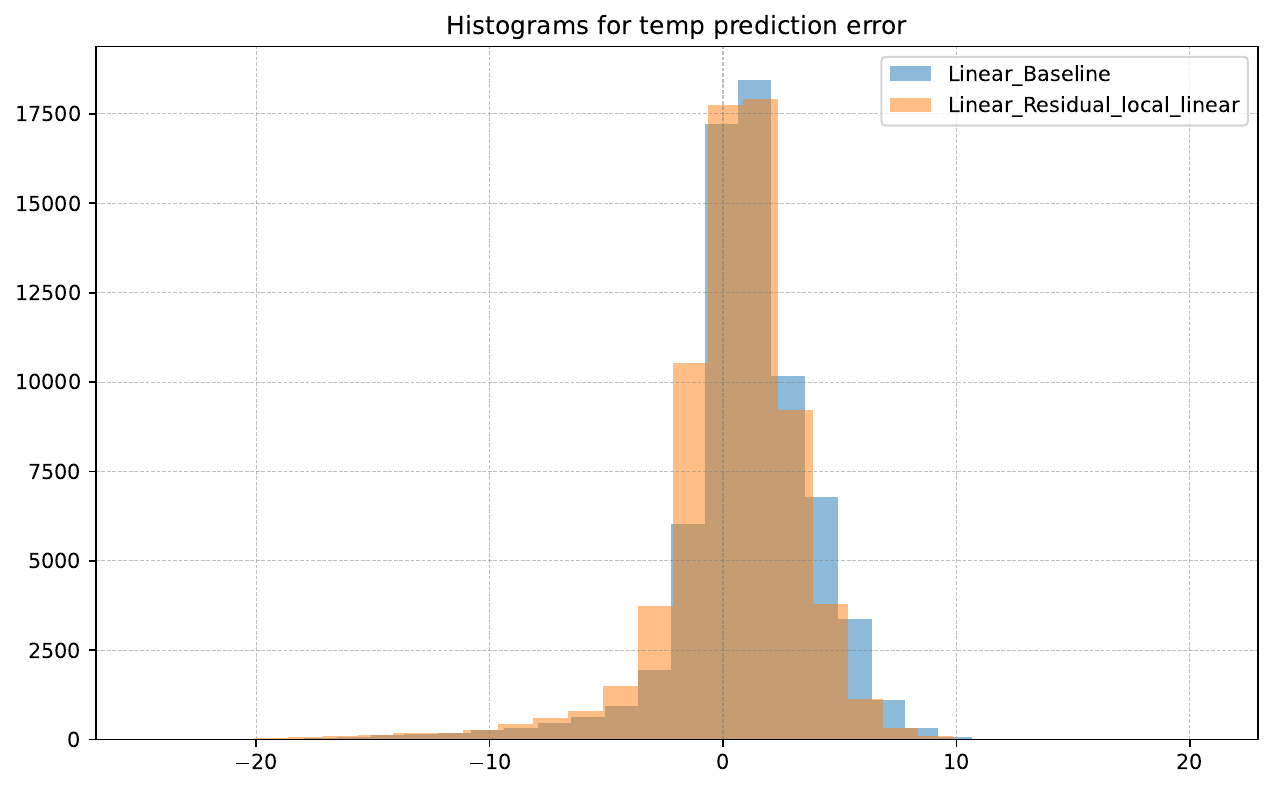}
  \caption{Histogram of temperature prediction error for the baseline model (\degC) and RCM with local forecasting strategy and linear weights for the subset 202204.}
  \label{fig:err_hist}
\end{figure}

\end{document}